\documentclass[11pt,a4paper]{article}
\usepackage[letterpaper]{geometry}
\usepackage[hyperref]{acl2020}
\usepackage{times}
\usepackage{latexsym}

\usepackage[section]{placeins}
\usepackage{subcaption}
\usepackage{amssymb}
\usepackage{amsmath}
\usepackage{todonotes}%
\usepackage{tikz}
\usetikzlibrary{arrows.meta}
\usepackage{tikz-qtree}
\usepackage{enumitem}
\usepackage{txfonts}  % makes equations narrower
\usepackage{stmaryrd}
\usepackage{multirow}
\usepackage{graphicx}
\usepackage{url}
\usepackage{mathtools}
\usepackage{comment}

\usepackage{microtype}

\aclfinalcopy % Uncomment this line for the final submission
\title{The Importance of Category Labels in Grammar Induction with Child-directed Utterances}

\author{Lifeng Jin \and William Schuler \\
        Department of Linguistics  \\
        The Ohio State University, Columbus, OH, USA\\
        {\tt \{jin, schuler\}@ling.osu.edu }
        }

\date{}

\def\GV{\mathbf{G}}% grammar
\def\NV{N}

\def\gl{1} % Gorn left
\def\gr{2} % Gorn right

\begin{document}
\maketitle
\begin{abstract}
Recent progress in grammar induction has shown that grammar induction is possible without explicit assumptions of language-specific knowledge. However, evaluation of induced grammars usually has ignored phrasal labels, an essential part of a grammar. Experiments in this work using a labeled evaluation metric, RH, show that linguistically motivated predictions about grammar sparsity and use of categories can only be revealed through labeled evaluation. Furthermore, depth-bounding as an implementation of human memory constraints in grammar inducers is still effective with labeled evaluation on multilingual transcribed child-directed utterances.
\end{abstract}

\section{\label{sec:intro}Introduction}

Recent work in probabilistic context-free grammar (PCFG) induction has shown that it is possible to learn accurate grammars from raw text \cite{Jin2018tacl,Jin2019flow,Kim2019a}, which is significant in addressing the issue of the \textit{poverty of the stimulus} \cite{Chomsky1965,chomsky1980poverty} in linguistics. 
Although 
phrasal categories and morphosyntactic features can be induced from raw text \cite{Jin2019vas,Jin2019flow}, most unsupervised parsing work has been evaluated using unlabeled parsing accuracy scores \cite{Seginer2007acl,Ponvert2011,Jin2018tacl,Shen2018,Shen2018bb,Shi2019}.
This is potentially distortative because
children and adults can distinguish categories of phrases and clauses \cite{Tomasello1993,Valian1986,Kemp2005syncat,Pine2013}, and much of acquisition modeling research has been directed at simulating the development of abstract linguistic categories in first language acquisition \cite{Bannard2009,Perfors2011,Kwiatkowski,Abend2017a,Jin2018tacl}.

Recent work proposed a labeled parsing accuracy evaluation metric called Recall-V-Measure (RVM) as a method for evaluating unsupervised grammar inducers \cite{Jin2019flow}, but this metric counts categories as incorrect if they are finer-grained than reference categories or if they represent binarizations of n-ary branches in reference trees, which may be linguistically acceptable. We therefore further modify it to Recall-Homogeneity (RH) calculated as the homogeneity \cite{rosenberghirschberg07} of the labels of matching constituents of the induced and gold trees, weighted by unlabeled recall.
This work uses transcribed child-directed utterances from multiple languages as input to a grammar inducer with hyperparameters tuned using either unlabeled F1 or labeled RH. Results show that: (1) the induced grammars capture the preference of sparse concentrations in human grammars only when using labeled evaluation; (2) grammar accuracy increases as the number of labels grows only when using labeled evaluation; (3) depth-bounding \cite[][ limiting center embedding]{Jin2018emnlp} is still effective when tuned to maximize labeled parsing accuracy.

\section{Model}

\begin{figure*}
\centering
\begin{subfigure}{.32\textwidth}
  \centering
  \includegraphics[width=\linewidth]{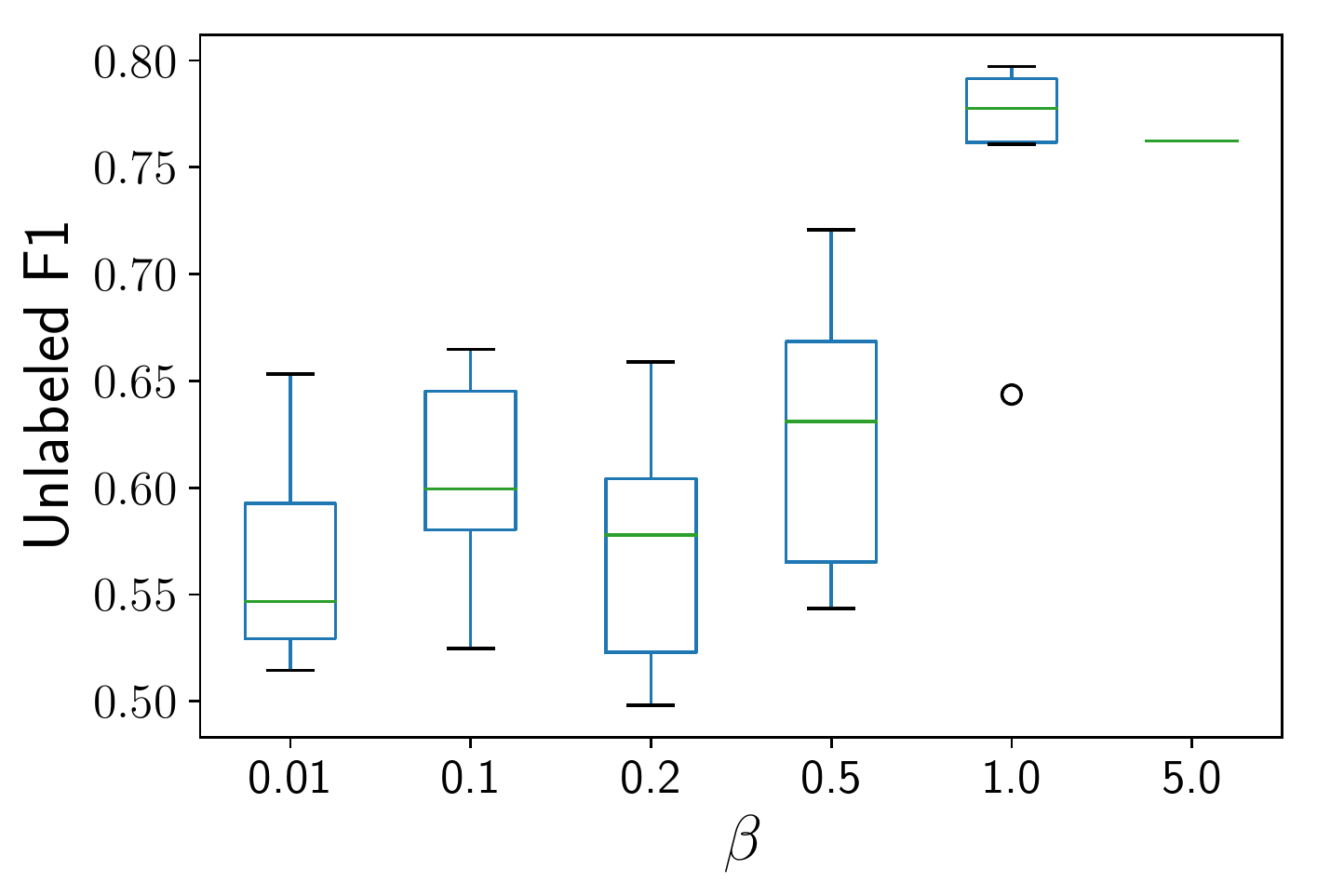}
  \caption{Unlabeled F1 scores with $\beta$s}
  \label{adamdev:sub1}
\end{subfigure}%
\begin{subfigure}{.32\textwidth}
  \centering
  \includegraphics[width=\linewidth]{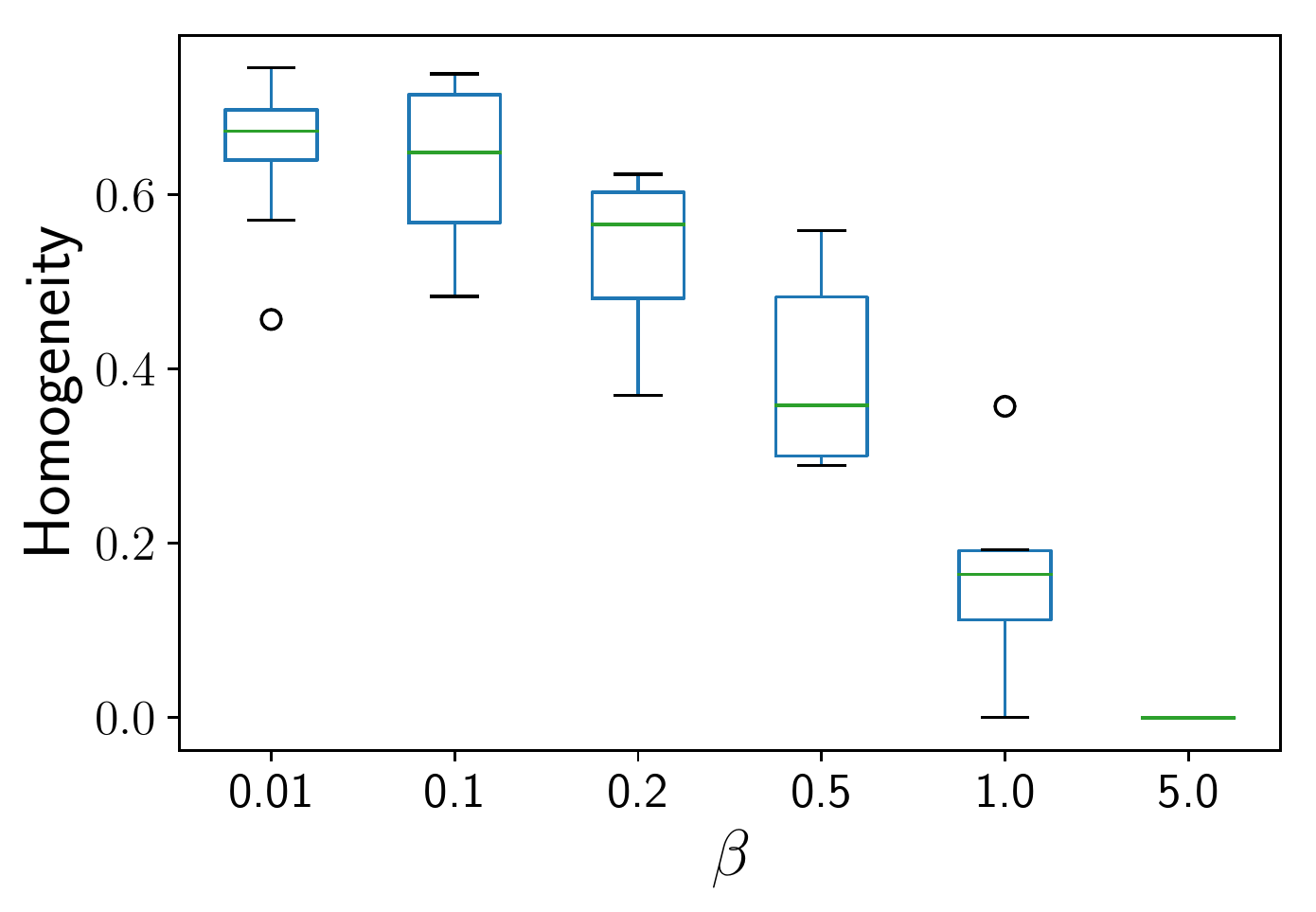}
  \caption{Homogeneity scores with $\beta$s}
  \label{adamdev:sub2}
\end{subfigure}
\begin{subfigure}{.32\textwidth}
  \centering
  \includegraphics[width=\linewidth]{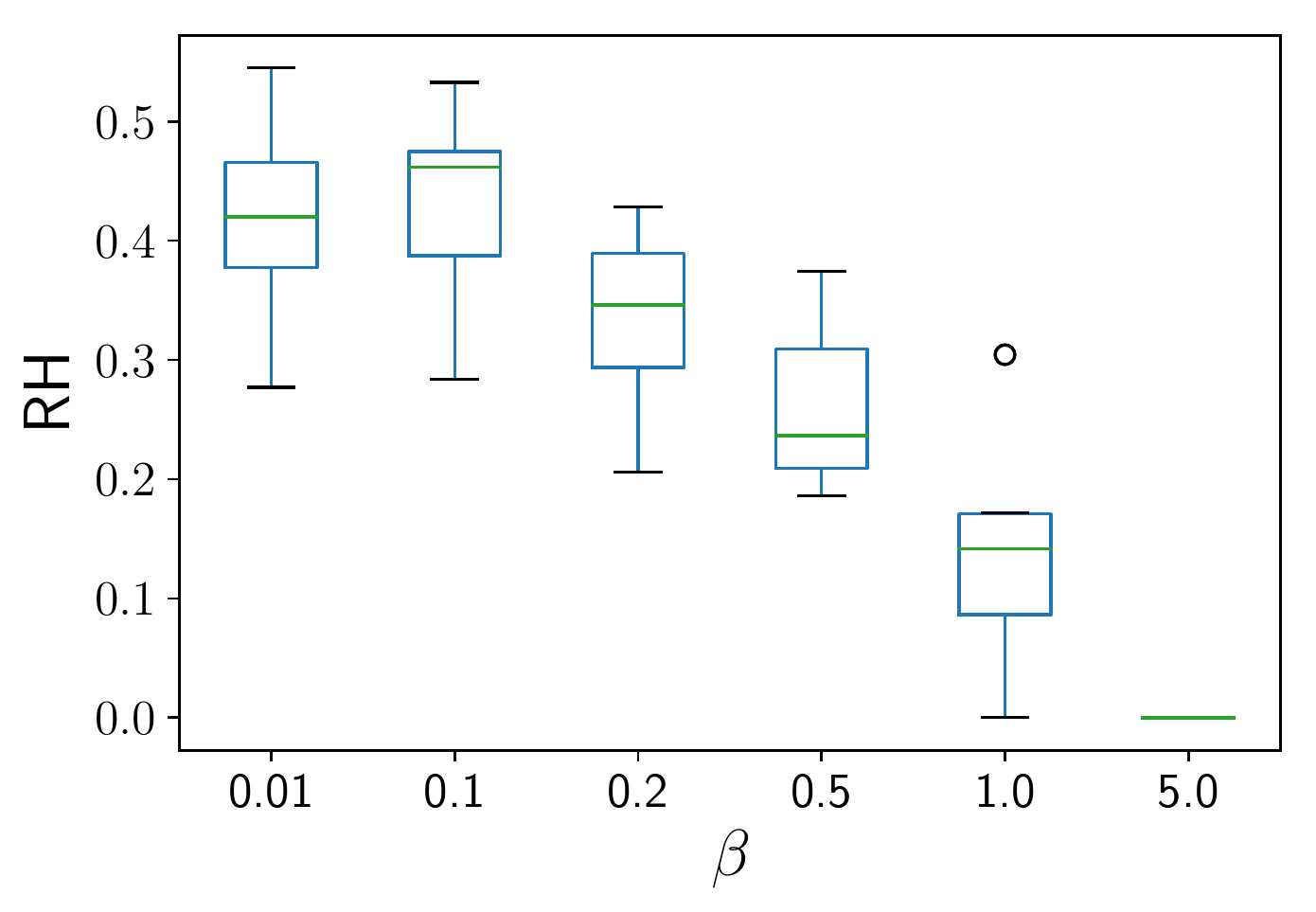}
  \caption{RH scores with $\beta$s}
  \label{adamdev:sub3}
\end{subfigure}
\begin{subfigure}{.32\textwidth}
  \centering
  \includegraphics[width=\linewidth]{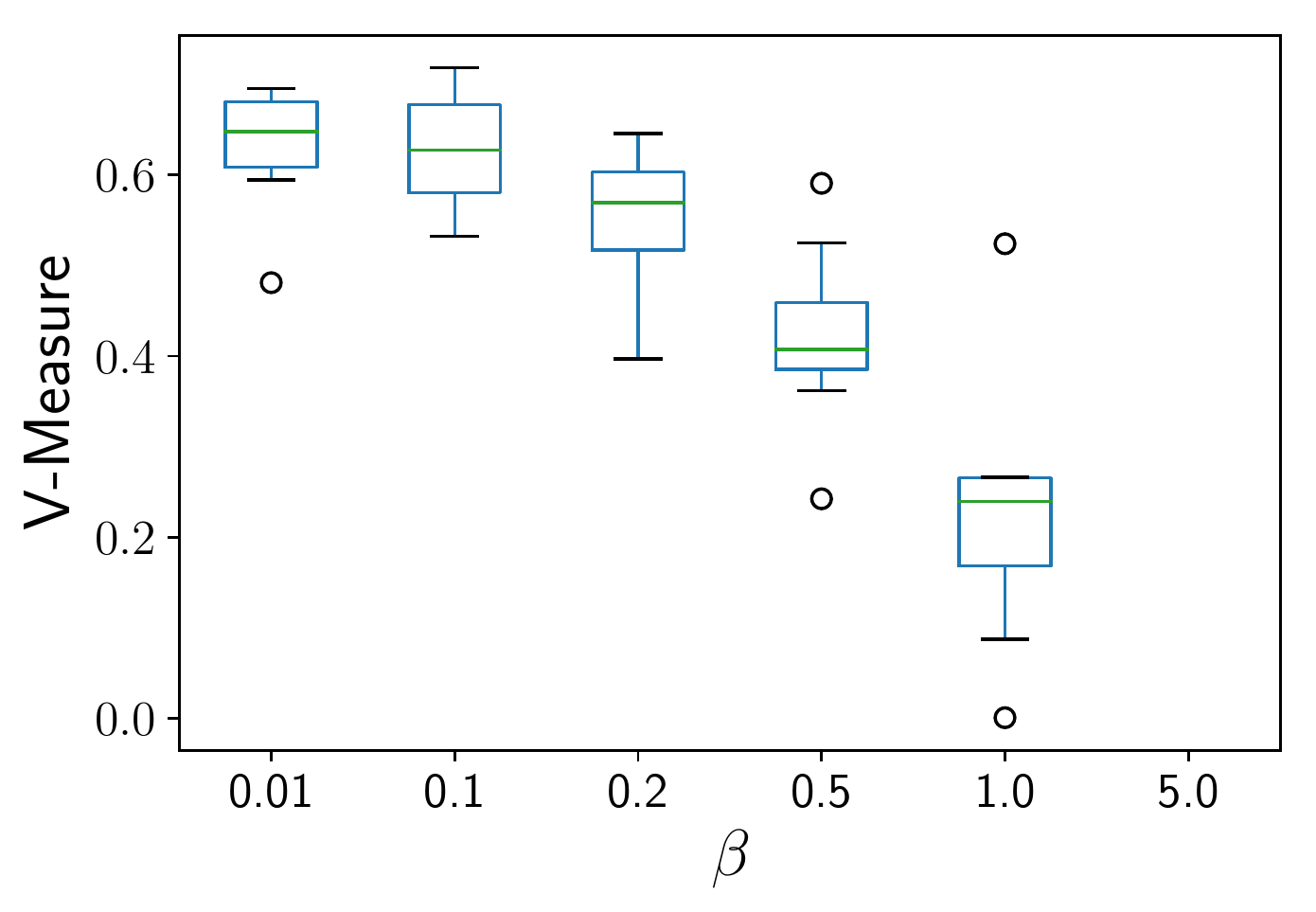}
  \caption{V-measure scores with $\beta$s}
  \label{adamdev:sub4}
\end{subfigure}
\begin{subfigure}{.32\textwidth}
  \centering
  \includegraphics[width=\linewidth]{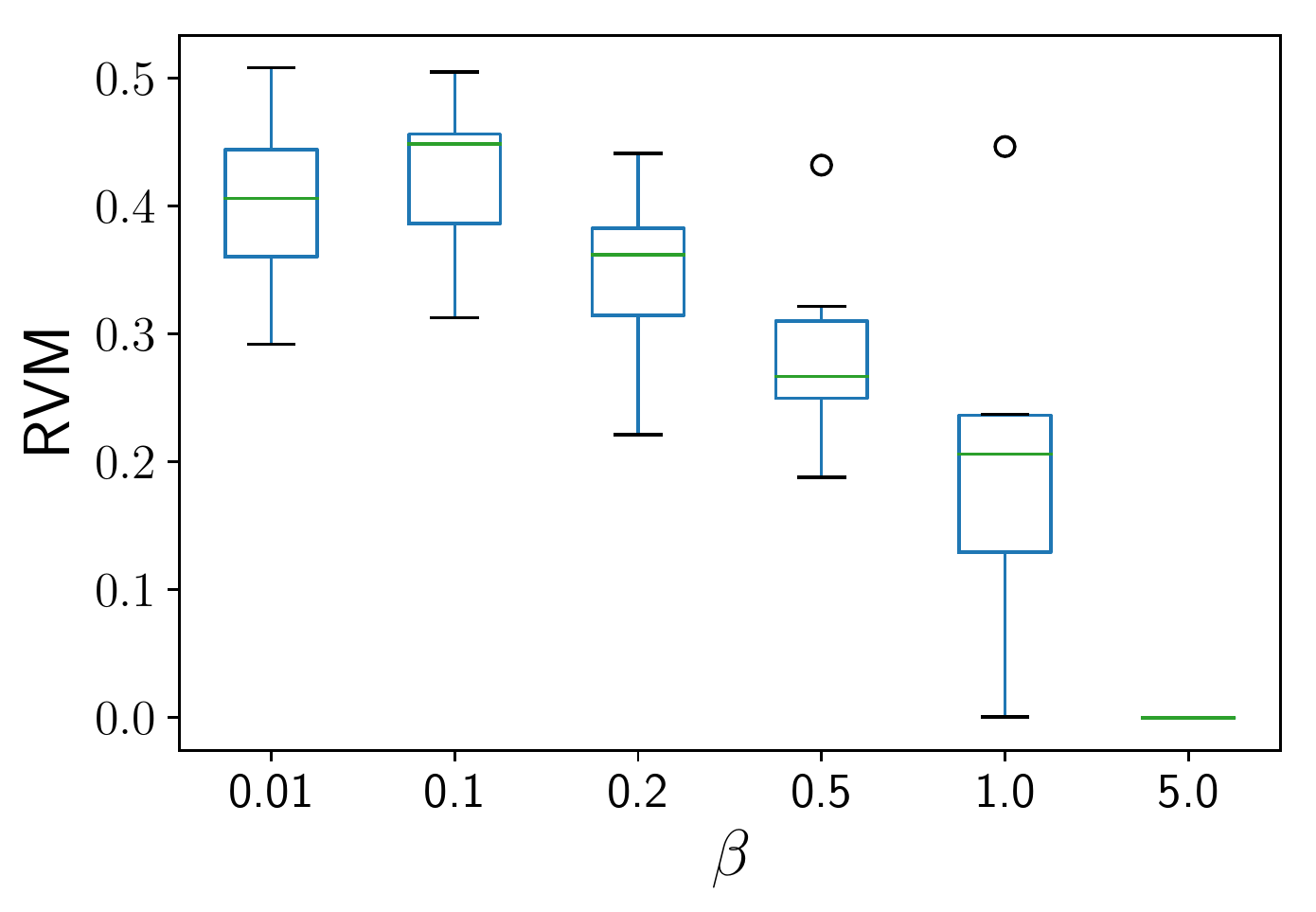}
  \caption{RVM scores with $\beta$s}
  \label{adamdev:sub5}
\end{subfigure}
\caption{Different evaluation metrics on the Adam dataset with different $\beta$ values.}
\label{fig:dev}
\end{figure*}

All experiments described in this paper use a Bayesian Dirichlet-multinomial model \cite{Jin2018emnlp} to induce PCFGs without assuming any language specific knowledge.
This model defines a Chomsky normal form (CNF) PCFG with $C$ nonterminal categories as a matrix~$\GV$ of binary rule probabilities
which is first drawn from the Dirichlet prior with a concentration parameter $\beta$:
\begin{equation}
\GV \sim \mathrm{Dirichlet}( \beta )
\end{equation}
Trees for sentences~$1..\NV$ in a corpus are then drawn from a PCFG parameterized by $\GV$:
\begin{equation}
\tau_{1..\NV} \sim \mathrm{PCFG}( \GV ),
\end{equation}
and each tree~$\tau$ is a set $\{ \tau_\epsilon, \tau_\gl, \tau_\gr, \tau_{\gl\gl}, \tau_{\gl\gr}, \tau_{\gr\gl}, ...\}$ of category node labels~$\tau_\eta$ where~$\eta \in \{\gl,\gr\}^*$ defines a path of left or right branches from the root to that node.
Category labels for every pair of left and right children~$\tau_{\eta \gl}, \tau_{\eta \gr}$ are drawn from a multinomial distribution defined by the grammar~$\GV$ and the category of the parent~$\tau_\eta$:
\begin{equation}
\tau_{\eta \gl}, \tau_{\eta \gr} \sim \mathrm{Multinomial}( {\delta_{\tau_\eta}\!}^\top \GV )
\end{equation}
where
$\delta_x$ is a Kronecker delta function equal to $1$ at value~$x$ and $0$ elsewhere. Terminal expansions are treated as expanding into a terminal node followed by a special null node.
% }

Inference in this model uses Gibbs sampling to produce samples of grammars and trees with the most probable parses obtained with the Viterbi algorithm.

\section{Data and hyperparameters}
Experiments here use transcribed child-directed utterances from the CHILDES corpus \cite{macwhinney00} in three languages with more than 15,000 sentences each. English hand-annotated constituency trees are taken from the Adam and Eve portions of the Brown Corpus \cite{brown1973first}. Mandarin \cite[Tong,][]{Deng2018Tong} and German \cite[Leo,][]{behrens2006input} data are collected from CHILDES with reference trees automatically generated using the state-of-the-art \citet{Kitaev} parser. Disfluencies are removed, and only sentences spoken by caregivers are kept in the data. Models are run 10 times with 700 iterations with random seeds following previous work \cite{Jin2018emnlp}. The last sampled grammar is used to generate Viterbi parses for all sentences. %
All punctuation is retained during induction and then removed in evaluation. Significance testing uses permutation tests on concatenations of Viterbi trees from all test runs. We use Adam for exploratory experiments and the other three sets for confirmatory experiments.
\begin{figure}
\centering
\begin{subfigure}{.4\textwidth}
  \centering
  \includegraphics[width=\linewidth]{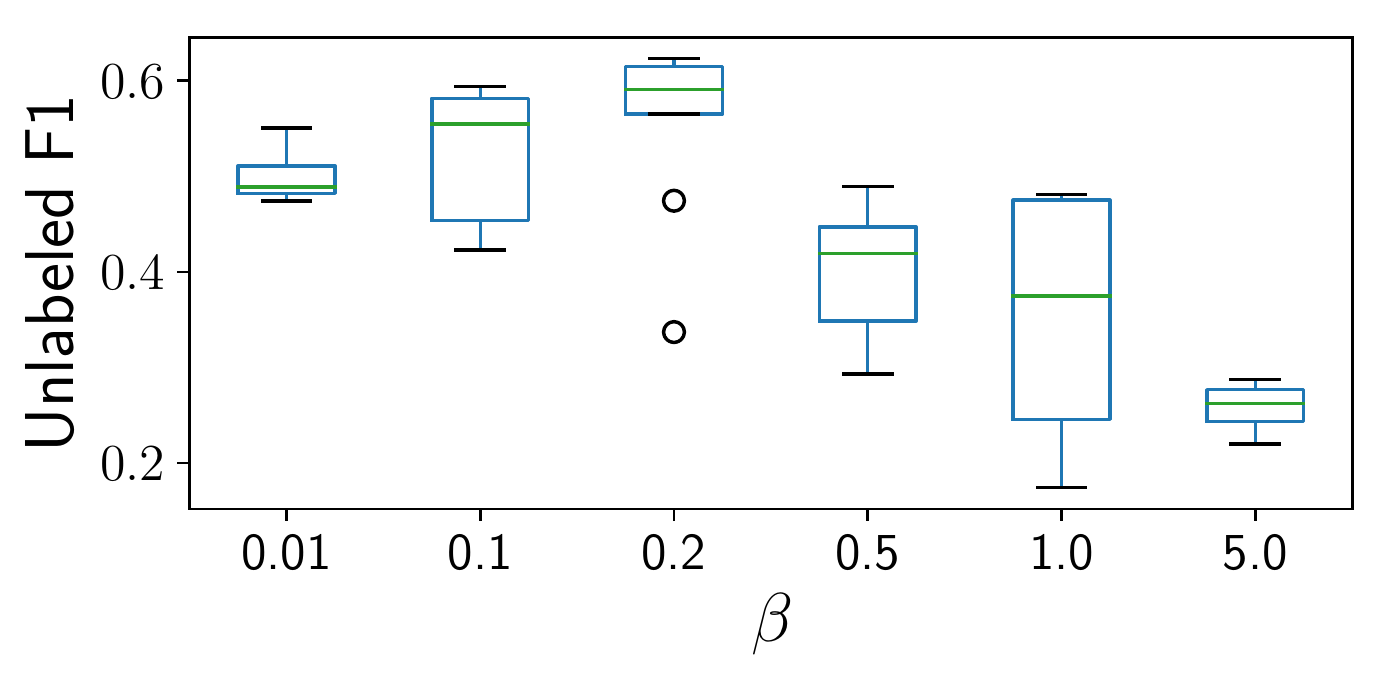}
  \caption{Unlabeled F1 scores with $\beta$s}
  \label{wsj20:sub1}
\end{subfigure}
\begin{subfigure}{.4\textwidth}
  \centering
  \includegraphics[width=\linewidth]{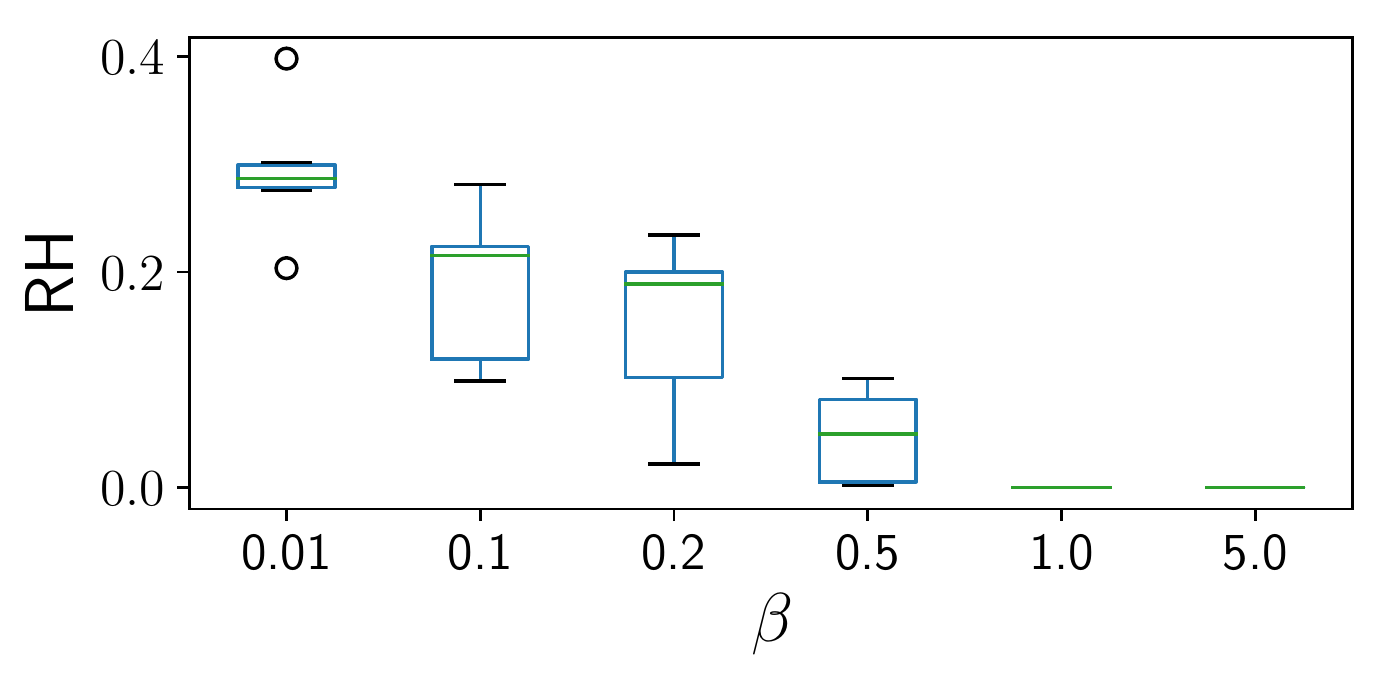}
  \caption{RH scores with $\beta$s}
  \label{wsj20:sub2}
\end{subfigure}
\caption{Different evaluation metrics on the WSJ20Dev dataset with different $\beta$ values.}
\label{fig:wsj20}
\end{figure}

\begin{figure*}
\centering
\begin{subfigure}{.32\textwidth}
  \centering
  \includegraphics[width=\linewidth]{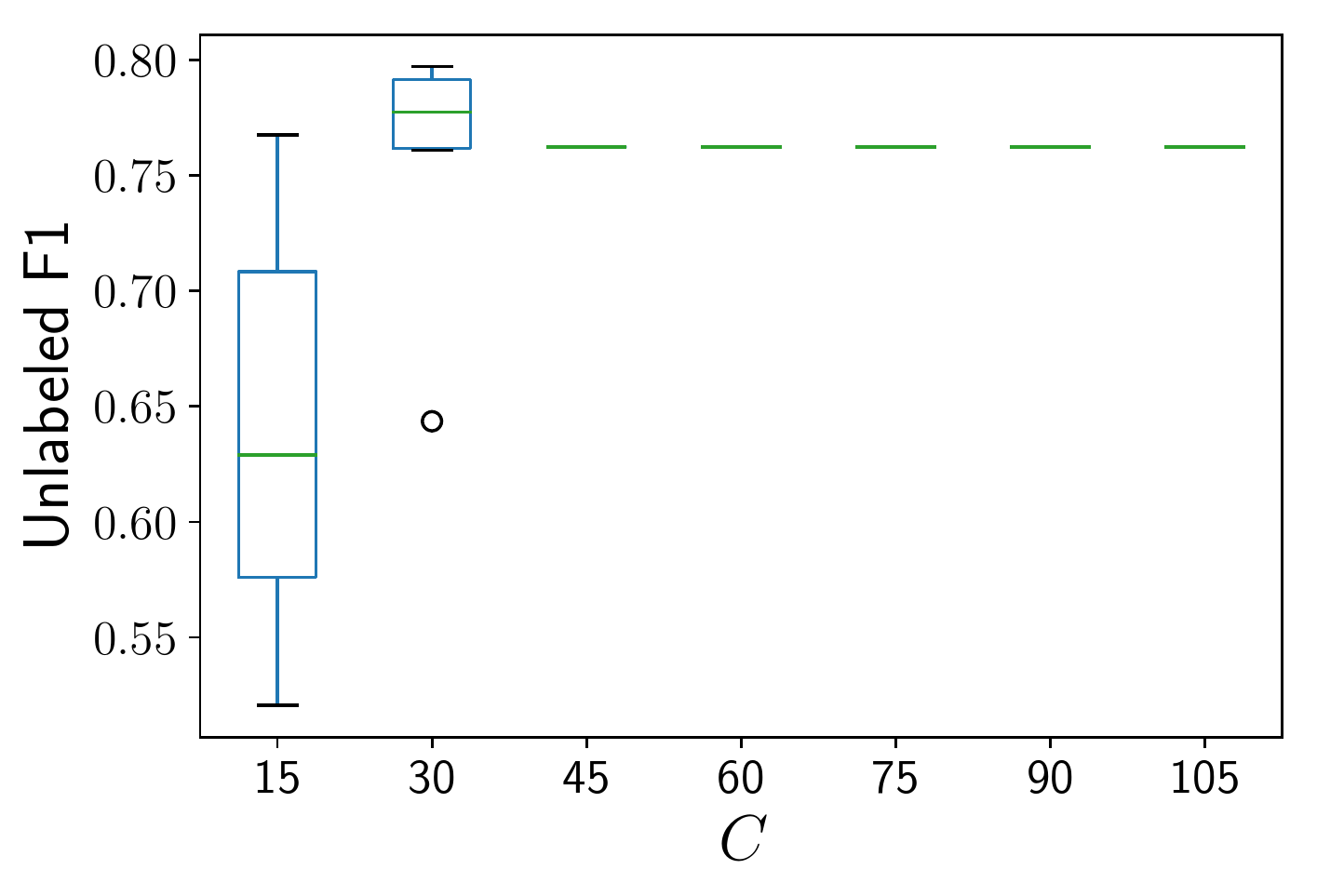}
  \caption{Unlabeled F1 with $C$s and $\beta=1$}
  \label{k:sub1}
\end{subfigure}%
\begin{subfigure}{.32\textwidth}
  \centering
  \includegraphics[width=\linewidth]{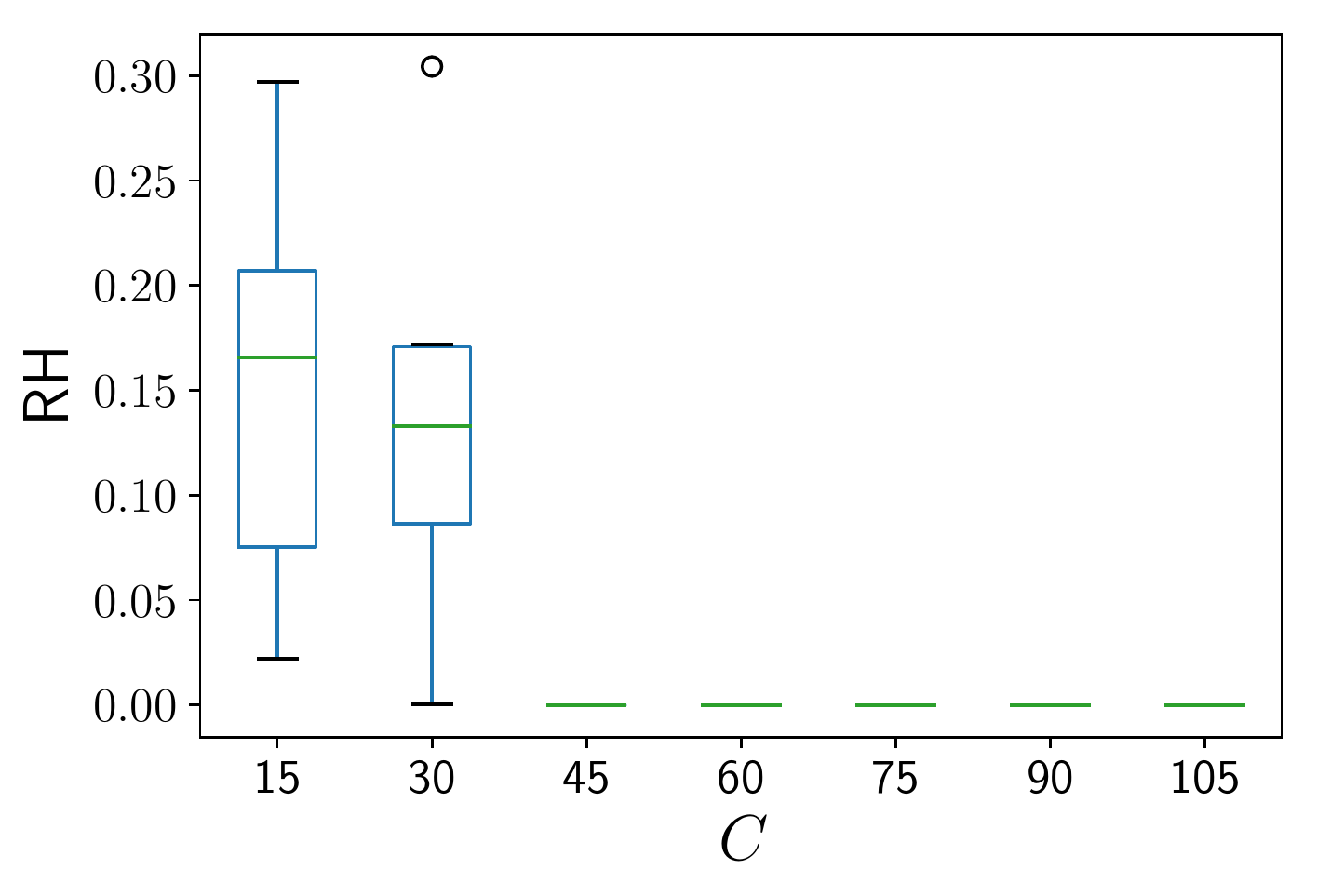}
  \caption{RH scores with $C$s and $\beta=1$}
  \label{k:sub2}
\end{subfigure}
\begin{subfigure}{.32\textwidth}
  \centering
  \includegraphics[width=\linewidth]{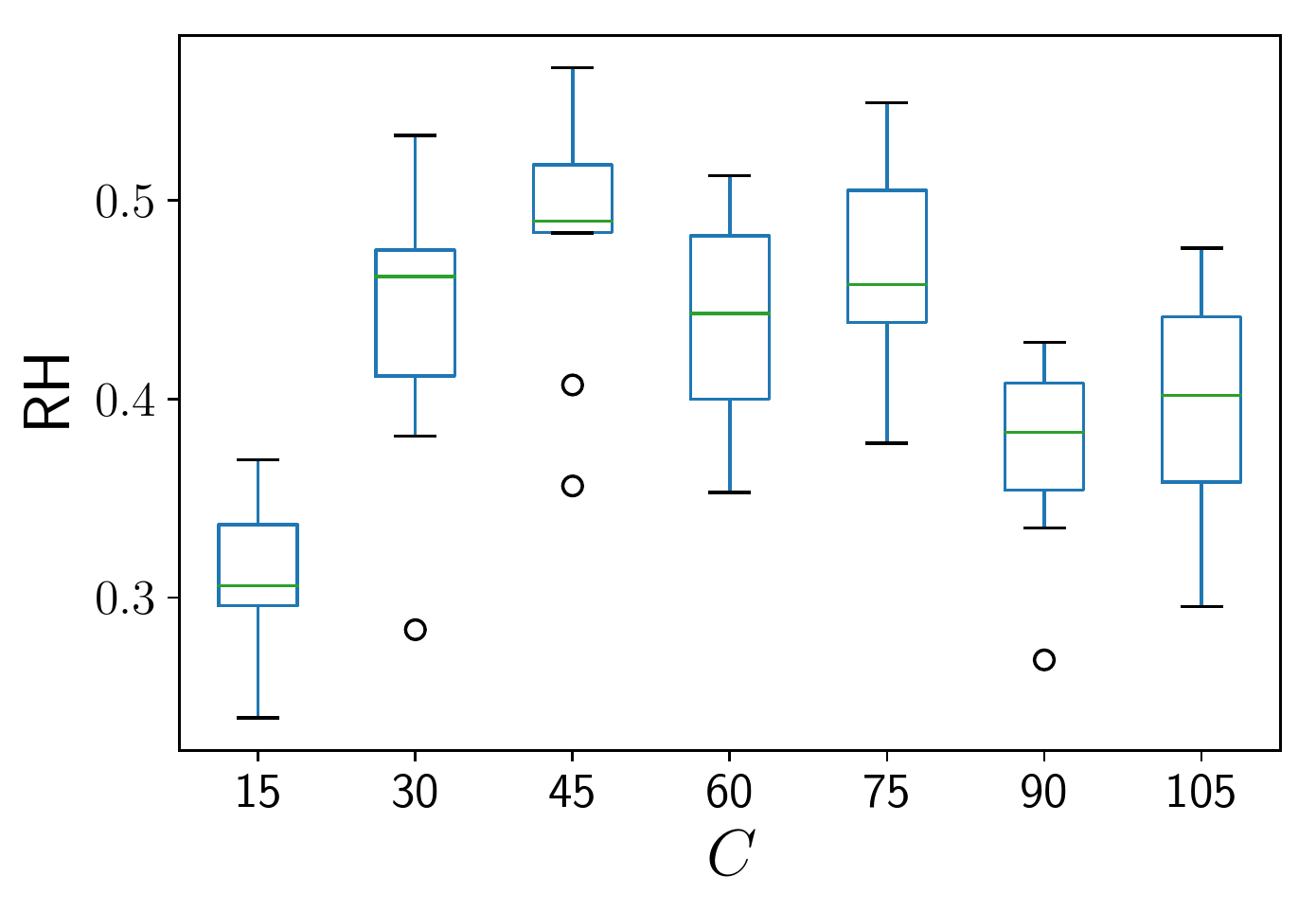}
  \caption{RH scores with $C$s and $\beta=0.01$}
  \label{k:sub3}
\end{subfigure}
\caption{Different evaluation metrics on the Adam dataset with different $C$ values at high and low $\beta$s.}
\label{fig:k}
\end{figure*}
\begin{figure}
\centering
  \includegraphics[width=.4\textwidth]{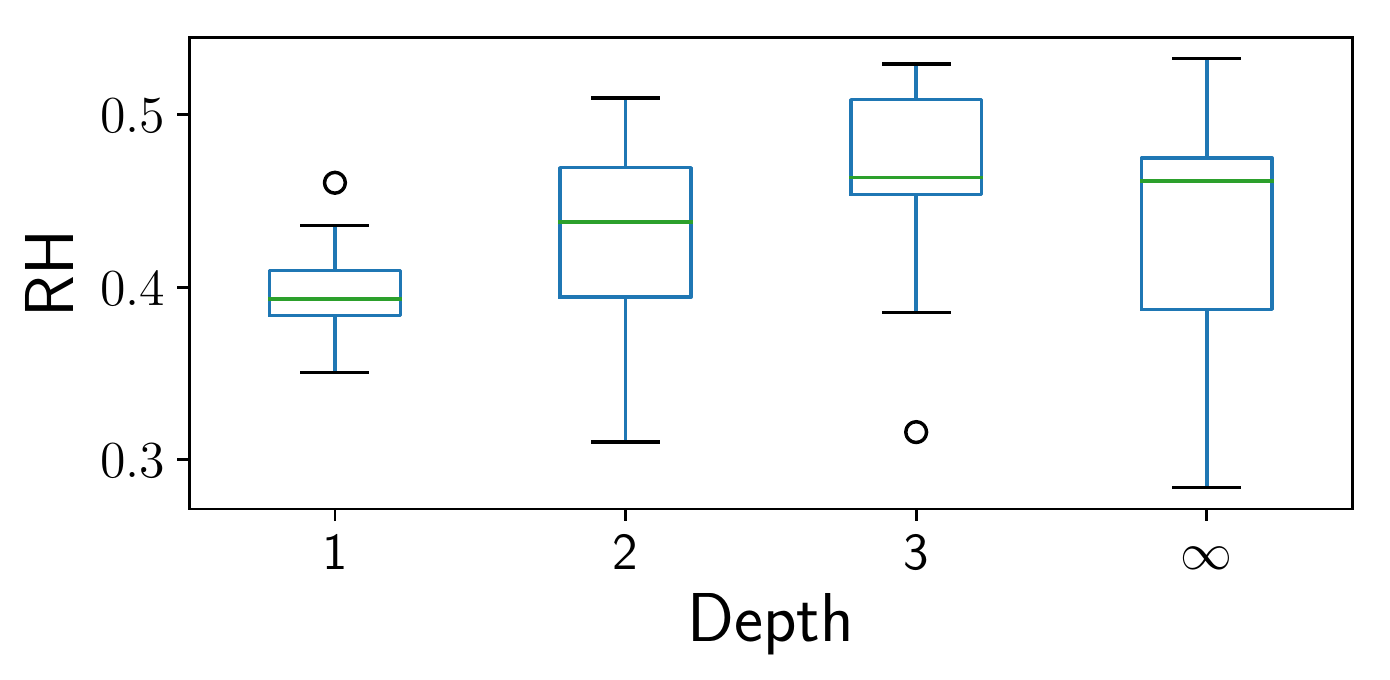}
  \caption{Depth-bounding on Adam}
  \label{adam:db}
\end{figure}

\subsection{Recall-Homogeneity}

RH is calculated by multiplying unlabeled recall of bracketed spans in the predicted Viterbi trees with the homogeneity score \cite{rosenberghirschberg07} of the predicted labels of the matching spans, This is different from RVM \citep{Jin2019flow}, which is the product of unlabeled recall and V-measure. The metric is insensitive to the branching factor of the grammar by the use of unlabeled recall. Unlike RVM, it is also insensitive to the precision of predicted labels to gold labels, indicating that models are not penalized by hypothesizing more refined categories, as long as these categories all fall into the confines of a gold category. RVM, on the other hand, would penalize both underproposing and overproposing categories compared to the ones in the annotation, but the gold categories, like nouns and verbs, are defined on a very high level that languages almost always further specify, represented usually as subcategories or features in linguistic theories.  
Unary branches in gold and predicted trees are removed, and the top category is used as the category for the constituent.

\section{Experiments}
\subsection{Experiment 1: Labeled evaluation shows preference of grammar sparsity}

Human grammars are sparse \cite{Johnson2007pcfg,Griffiths2007}. For example, in the Penn Treebank \cite{marcusetal93}, there are 73 unique nonterminal categories. In theory, there can be more than 28 million possible unary, binary and ternary branching rules in the grammar. However, only 17,020 unique rules are found in the corpus, showing the high sparsity of attested rules. In other frameworks like Combinatory Categorial Grammar \cite{Steedman2002} where lexical categories can be in the thousands, the number of attested lexical categories is still small compared to all possible ones.

The Dirichlet concentration hyperparameter $\beta$ in the model controls the probability of a sampled multinomial distribution concentrating its probability mass on only a few items. Previous work using similar models usually sets this value low \cite{Johnson2007pcfg,Griffiths2007,Graca,Jin2018tacl} to prefer sparse grammars (i.e.\ grammars in which most of the probability mass is allocated to a small number of rules), with good results. The prediction based on the preference of sparsity is that the best $\beta$ value should be much lower than 1.

Figure \ref{adamdev:sub1} shows unlabeled F1 scores with different $\beta$ values on Adam.%
\footnote{The results shown in the figure use $C$=30. We also tested other $C$ values from 15 to 105 and the trend is almost identical.}
Contrary to the prediction, grammar accuracy peaks at high values for $\beta$ when measured using unlabeled F1. 
However, these grammars with high unlabeled F1 are almost purely right-branching grammars, which performs very well on English child-directed speech in unlabeled parsing evaluation, but the right-branching grammars have phrasal labels that do not correlate with human annotation when evaluated with Homogeneity, shown in Figure \ref{adamdev:sub2}. This indicates that instead of capturing human intuitions about syntactic structure, such grammars have only captured broad branching tendencies.
The same grammars are evaluated again with RH, shown in Figure \ref{adamdev:sub3}. When both structural and labeling accuracy is taken into account, results correctly capture the intuition that grammar accuracy has a low peaking concentration hyperparameter. Figure \ref{adamdev:sub4} and \ref{adamdev:sub5} shows the same experiments evaluated with the labeled evaluation metric RVM. Because of the sensitivity to labeling accuracy, results in VM and RVM also show the similar trend as Homogeneity and RH where labeling quality decreases as $\beta$ increases. \citet{Jin2018tacl} noted that induced grammars high in unlabeled bracketing scores are low in NP discovery scores, which is a category-specific evaluation metric. This can also be explained by the induced grammars with high bracketing scores only capture a broad right-branching bias without accurately clustering words and phrases based on their distributional properties.

Figure \ref{fig:wsj20} shows the same experiments on a corpus of formal English written text, the WSJ20dev%
\footnote{The first half of the Wall Street Journal part of the Penn Treebank with sentences with 20 words or fewer.}
dataset.
The pattern is similar but less extreme than on CHILDES.
The higher $\beta$s at the range of 0.1-0.2 still show better performance on unlabeled F1 than the sparser models, consistent with previous results in \citet{Jin2018tacl}. However RH scores reveal that the labels induced by the denser models are less accurate, manifesting as the overall lower peak for $\beta$ using RH than using unlabeled F1.

\subsection{Experiment 2: Performance increases with the number of categories}

\begin{figure*}
\centering
\begin{subfigure}{.32\textwidth}
  \centering
  \includegraphics[width=\linewidth]{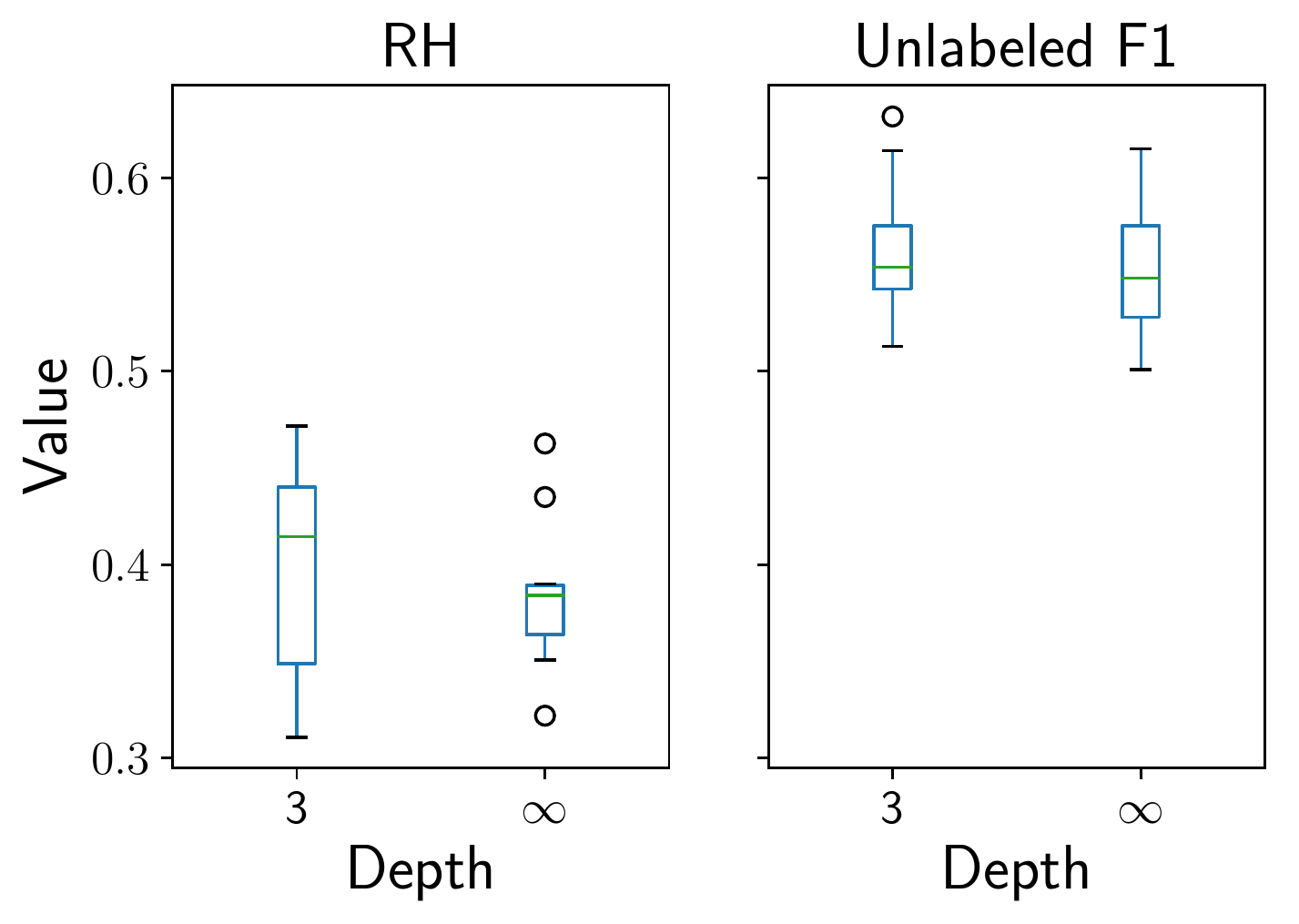}
  \caption{Depth-bounding on Eve}
  \label{db:sub1}
\end{subfigure}%
\begin{subfigure}{.32\textwidth}
  \centering
  \includegraphics[width=\linewidth]{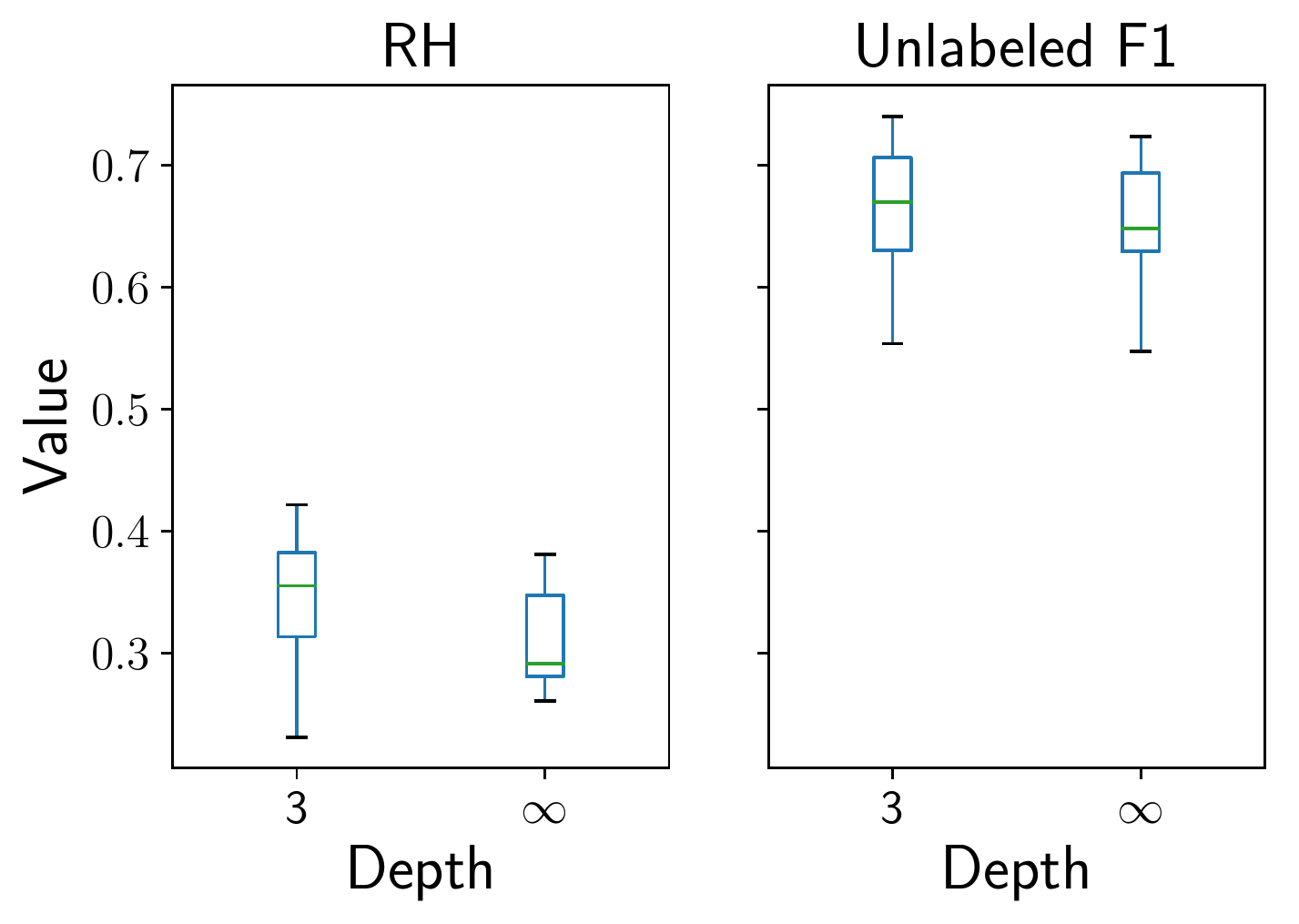}
  \caption{Depth-bounding on Tong}
  \label{db:sub2}
\end{subfigure}
\begin{subfigure}{.32\textwidth}
  \centering
  \includegraphics[width=\linewidth]{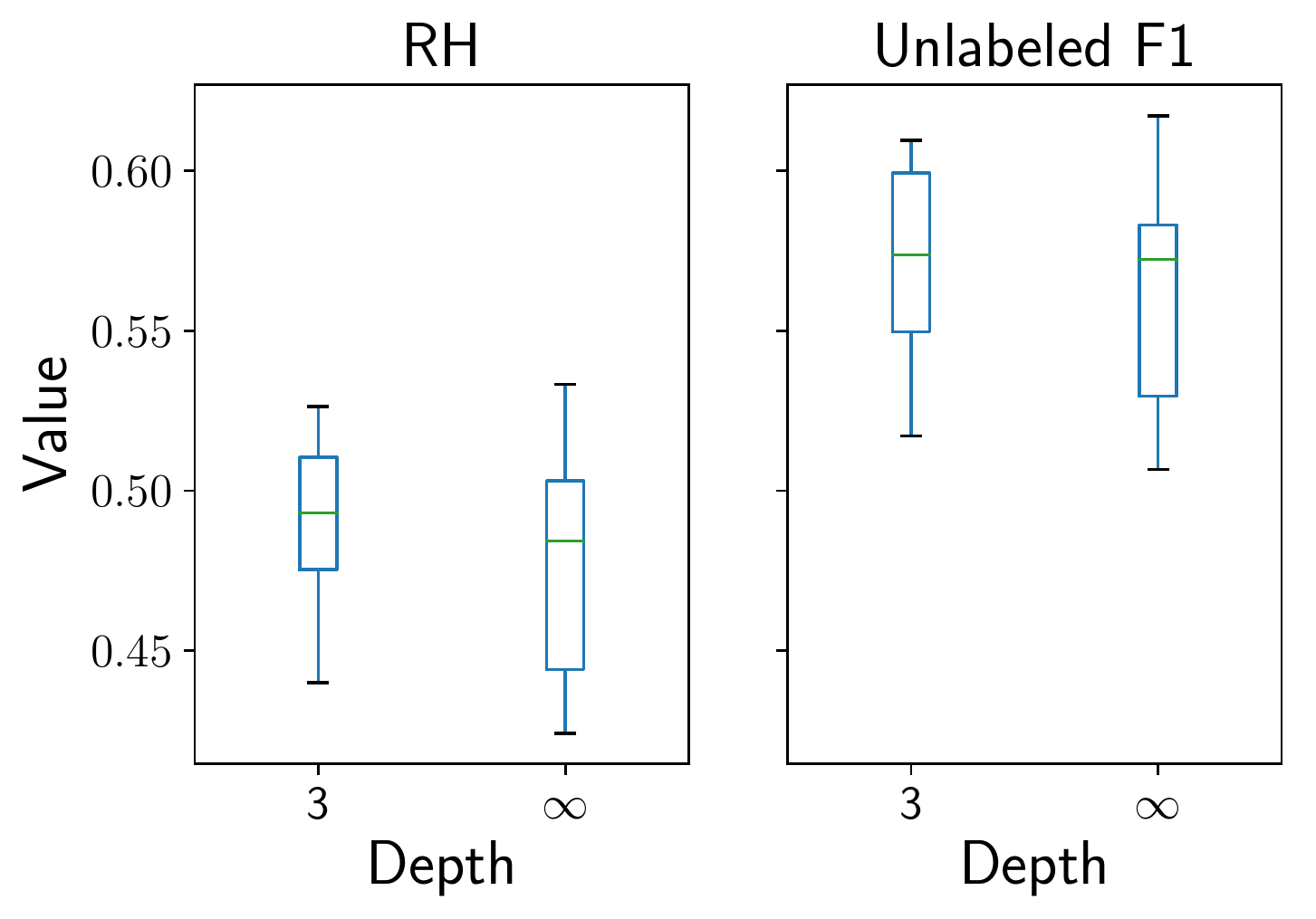}
  \caption{Depth-bounding on Leo}
  \label{db:sub3}
\end{subfigure}
\caption{Comparison of labeled and unlabeled evaluation of grammars bounded at depth 3 and unbounded grammars on English (Eve), Chinese Mandarin (Tong) and German (Leo) datasets from CHILDES.}
\label{fig:db}
\end{figure*}

Previous research \cite{Jin2018emnlp} also reported that the number of categories $C$ used by the induction models was relatively low compared to the number of categories in human annotation. For example, there are 63 unique tags in the Adam dataset. This is in contrast to 30 or fewer categories used in previous induction work. The bias brought by high $\beta$ values and unlabeled evaluation together may be masking the real relationship between the number of categories and grammar accuracy. 

Figures \ref{k:sub1} and \ref{k:sub2} show unlabeled and labeled evaluation on different grammars induced with the best performing $\beta$ on Adam tuned by unlabeled F1. With F1, increasing the number of categories beyond 30 yields no improvement as most of the induced grammars are purely right-branching grammars. RH results confirm this: as grammars approach the pure right-branching solution when $C$ increases, the similarity between induced and gold labels of constituents deteriorates quickly.
RH scores from grammars induced with $\beta=0.01$ are more indicative of the interaction between the number of categories and grammar accuracy. Grammar accuracy increases as $C$ gets larger initially and peaks at $C=75$. The results confirm the importance of labeled evaluation, because the trend from labeled evaluation shows that there should be a sufficient number of categories to account for different syntactic structures, and models with small numbers of categories are limited in their ability to do this.

\subsection{Experiment 3: Depth-bounding is still effective with RH}

Previous work showed that depth-bounding is effective in helping grammar inducers induce more accurate grammars \cite{Shain2016,Jin2018emnlp}, because it removes the parse trees with deeply nested center-embeddings, which cannot be produced by humans due to memory constraints \cite{chomskymiller63}, from grammar induction inference. However the unlabeled evaluation metric used in previous work may lead to unhelpful conclusions. In order to revisit this claim with labeled evaluation, experiments are first conducted on Adam exploring the interaction between depth and labeled performance, and subsequently on the Eve (English), Tong (Chinese Mandarin) and Leo (German) portions of the CHILDES corpus. All experiments use hyperparameters tuned with RH.%
\footnote{The optimal $C$ is 75 from previous experiments, but we used 30 in all depth-bounding experiments due to hardware constraints at high depth bounds.}

Figure \ref{adam:db} shows the interaction between depth and RH scores on Adam. Performance of the unbounded models can be lower than all bounded models, showing that unbounded inducers can induce grammars inconsistent with human memory constraints. 
The labeled performance peaks at depth 3, which is significantly more accurate ($p < 1 \times 10^{-3}$) than unbounded models. This is consistent with previous results that over 97\% of trees in English contain 3 or fewer nested center embeddings \cite{Schuler2010}.

Experiments on Eve, Tong and Leo replicate this result. Figure \ref{fig:db} shows that the models bounded at depth 3 are more accurate than unbounded models with both unlabeled and labeled evaluation metrics. Significance testing with unlabeled F1%
\footnote{Neither RH nor RVM were used in permutation significance testing, because labels with the same values from different induced grammars may represent different linguistic categories, therefore two parses of the same sentence from different runs are not exchangeable.}
shows the performance differences across three datasets are all highly significant ($p < 0.001$). Therefore, the claim that depth-bounding is effective in grammar induction is still supported when the models are developed and evaluated with labeled evaluation.

\section{Conclusion}

Unlabeled evaluation has been used in grammar induction, but experiments presented in this paper show that unlabeled evaluation can reveal unexpected bias in the data which may lead to unhelpful conclusions compared to labeled evaluation. Results show that trends of preference of sparsity and use of categories that are consistent with linguistic annotation can only be discovered with labeled evaluation. Furthermore, human memory constraints are still effective in grammar induction when labeled evaluation is used throughout all stages of development.

\FloatBarrier

\bibliography{references}
\bibliographystyle{acl_natbib}

\end{document}